\makeatletter\def\graphicscache@inhibit{true}\makeatother
\documentclass[a4paper,twoside]{article}

\usepackage[hidelinks]{hyperref}
\usepackage{booktabs}
\usepackage{epsfig}
\usepackage{subcaption}
\usepackage{calc}
\usepackage{amssymb}
\usepackage{amstext}
\usepackage{amsmath}
\usepackage{amsthm}
\usepackage{multicol}
\usepackage{pslatex}
\usepackage{apalike}
\usepackage[capitalize]{cleveref}

\usepackage[draft,nomargin,inline]{fixme}
\fxsetup{theme=color}

\usepackage{graphicscache}

\usepackage{SCITEPRESS}     %

\begin{document}

\title{Visual Descriptor Learning from Monocular Video}

\author{
\authorname{Umashankar Deekshith \sup{1} \orcidAuthor{0000-0002-0341-5768}
, Nishit Gajjar \sup{1} \orcidAuthor{0000-0001-7610-797X}
,  Max Schwarz\sup{1} \orcidAuthor{0000-0002-9942-6604}
and Sven Behnke\sup{1} \orcidAuthor{0000-0002-5040-7525}}
\affiliation{\sup{1}Autonomous Intelligent Systems group of University of Bonn, Germany}
\email{s6umdeek@uni-bonn.de, nishit.gajjar@rwth-aachen.de}
}


\abstract{Correspondence estimation is one of the most widely researched and yet only partially solved area of computer vision with many applications in tracking, mapping, recognition of objects and environment. In this paper, we propose a novel way to estimate dense correspondence on an RGB image where visual descriptors are learned from video examples by training a fully convolutional network. Most deep learning methods solve this by training the network with a large set of expensive labeled data or perform labeling through strong 3D generative models using RGB-D videos. Our method learns from RGB videos using contrastive loss, where relative labeling is estimated from optical flow. We demonstrate the functionality in a quantitative analysis on rendered videos, where ground truth information is available. Not only does the method perform well on test data with the same background, it also generalizes to situations with a new background. The descriptors learned are unique and the representations determined by the network are global. We further show the applicability of the method to real-world videos.}

\onecolumn \maketitle \normalsize \setcounter{footnote}{0} \vfill

\section{\uppercase{Introduction}}
\label{sec:introduction}

\begin{figure}[t]
	\begin{center}
		\includegraphics[width=\linewidth]{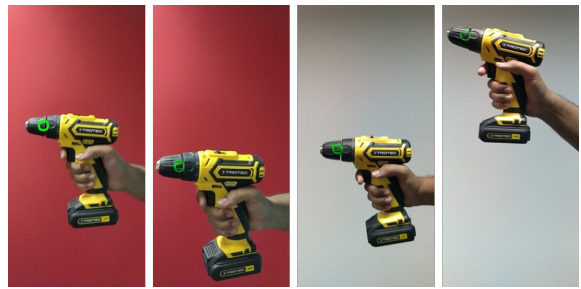}
	\end{center}
	\caption{Point tracking using the learned descriptors in monocular video.
	Two different backgrounds are shown to represent the network's capability to generate global correspondences. A square patch of 25 pixels are selected (left) and the nearest neighbor for those pixels based on the pixel representation is shown for the other images.}
	\label{fig:top_right}
\end{figure}

Many of the problems in computer vision, like 3D reconstruction, visual odometry, simultaneous localization and mapping (SLAM), object recognition, depend on the underlying problem of image correspondence (see \cref{fig:top_right}). Correspondence methods based on sparse visual descriptors like SIFT~\cite{SIFT} have been shown to be useful for applications like camera calibration, panorama stitching and even robot localization. However, SIFT and similar hand designed features require a textured image. They perform poorly for images or scenes which lack sufficient texture. In such situations, dense feature extractors are better suited than sparse keypoint-based methods.

With the advancement in deep learning in recent years, the general trend is that neural networks can be trained to outperform hand designed feature methods for any function using sufficient training data. However, supervised training approaches require significant effort because they require labeled training data. Therefore, it is useful to have a way to train the model in a self-supervised fashion, where the training labels are created automatically.

\begin{figure*}
	\begin{center}
		\includegraphics[width=\linewidth]{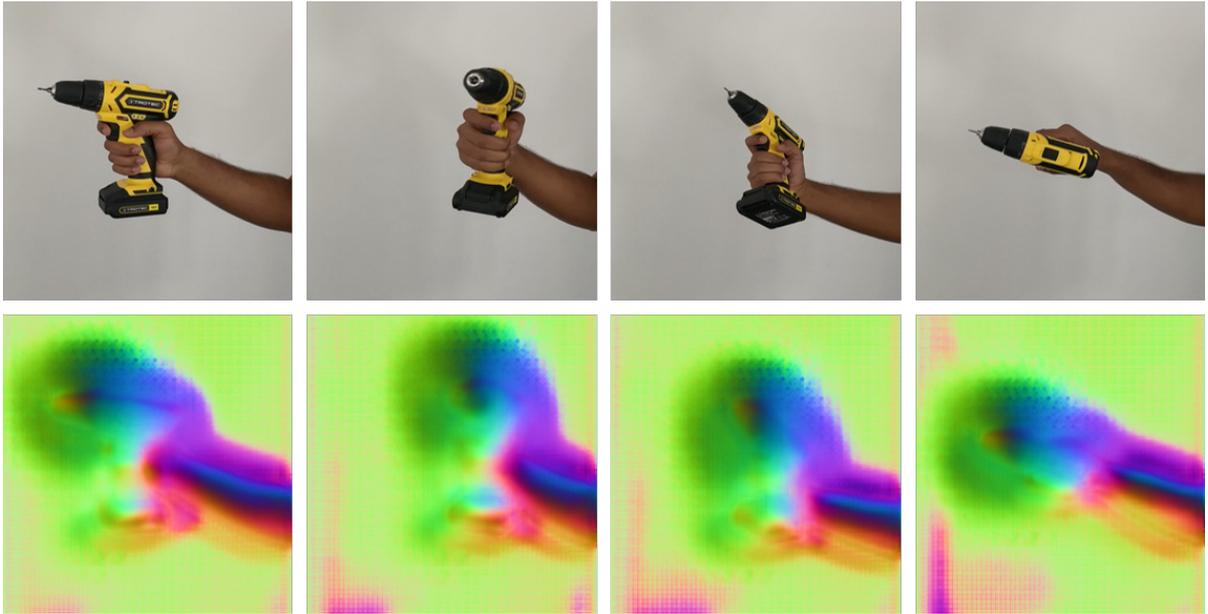}
	\end{center}
	\caption{Learned dense object descriptors. Top: Stills from a monocular video demonstrating full 6D movement of the object.
	Bottom: Normalized output of the trained network where each pixel is represented uniquely, visualized as an RGB image.
	The objects were captured in different places under different lighting conditions, viewpoints giving the object a geometric transformation in a 3D space. It can be seen here that the descriptor generated is independent of these effects.
	This video sequence has not been seen during training.}
	\label{fig:Opening_picture}
\end{figure*}

In their work, \cite{Schmidt_Sel} and \cite{Florence_Den} have shown approaches for self-supervised visual descriptor learning using raw RGB-D sequence of images. Their approach shows that it is possible to generate dense descriptors for an object or a complete scene and that these descriptors are consistent across videos with different backgrounds and camera alignments.
The dense descriptors are learned using contrastive loss~\cite{Hadsell_Dim}. The method unlocks a huge potential for robot manipulation, navigation, and self learning of the environment. However, for obtaining the required correspondence information for training, the authors rely on 3D reconstruction from RGB-D sensors. This limits the applicability of their method (i.e. shiny and transparent objects cannot be learned).

Every day a large quantity of videos are generated from basic point and shoot cameras. Our aim is to learn the visual descriptors from an RGB video without any additional information. We follow a similar approach to \cite{Schmidt_Sel} by implementing a self-supervised visual descriptor learning network which learns dense object descriptors (see \cref{fig:Opening_picture}). Instead of the depth map, we rely on the movement of the object of interest. To find an alternate way of self learning the pixel correspondences, we turn to available optical flow methods. The traditional dense optical flow method of \cite{Farnebaeck_opticalflow} and new deep learning based optical flow methods \cite{Flow_3,FlowNet2_p} provide the information which is the basis of our approach to generate self-supervised training data. Optical flow gives us a mapping of pixel correspondences within a sequence of images. Loosely speaking, our method turns \textit{relative} correspondence from optical flow into \textit{absolute} correspondence using our learned descriptors.

To focus the descriptor learning on the object of interest, we employ a generic foreground segmentation method~\cite{Chen_Deeplab}, which provides a foreground object mask. We use this foreground mask to limit the learning of meaningful visual descriptors for the foreground object, such that these descriptors are as far apart in descriptor space as possible.

In this work, we show that it is possible to learn visual descriptors for monocular images through self-supervised learning by training them using contrastive loss for images labeled from optical flow information.
We further demonstrate applicability in experiments on synthetic and real data.
\section{Related Work}

Traditionally, dense correspondence estimation algorithms were of two kinds. One with focus on learning generative models with strong priors \cite{Hinton_Wake}, \cite{Sudderth_des}. These algorithms were designed to capture similar occurrence of features. Another set of algorithms use hand-engineered methods. For example, SIFT or HOG performs clustering over training data to discover the feature classes \cite{Sivic_Disc}, \cite{Russel_Usin}.
Recently, the advances in deep learning and their ability to reliably capture high dimensional features directly from the data has made a lot of progress in correspondence estimation, outperforming the traditional methods. \cite{Taylor_Vit} have proposed a method where a regression forest is used to predict dense correspondence between image pixels and vertices of an articulated mesh model. Similarly, \cite{Shotton_Scene} use a random forest given only a single acquired image, to deduce the pose of an RGB-D camera with respect to a known 3D scene. \cite{Brachmann_Learn} jointly train an objective over both 3D object coordinates and object class labeling to determine to address the problem of estimating the 6D Pose of specific objects from a single RGB-D image.
Semantic segmentation of images as presented by \cite{Long_Fully,Hariharan_Hyper} use neural network that produce dense correspondence of images. \cite{Guler_Den} propose a method to establish dense correspondence between RGB image and surface based representation of human body. All of these methods rely on labeled data. To avoid the expensive labeling process, we use relative labels for pixels generated before training with minimum computation and no human intervention.

\begin{figure*}[t]
	\begin{center}
\includegraphics[width=\linewidth]{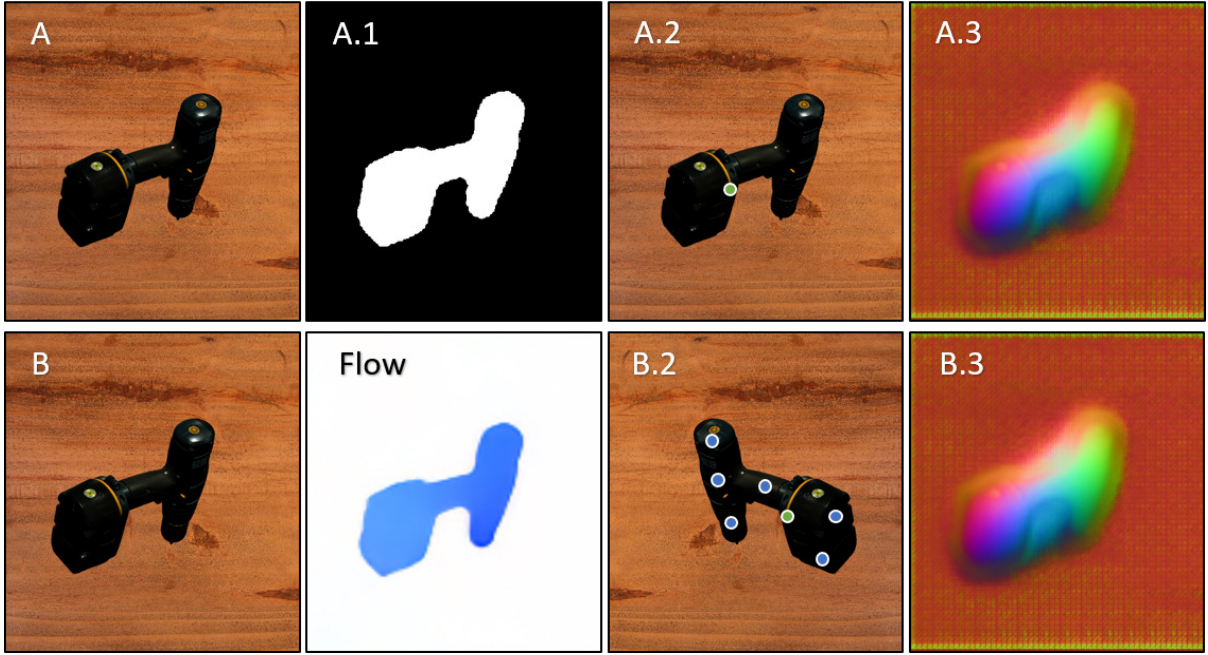}
	\end{center}
	\caption{Training process.
	Images \textit{A} and \textit{B} are consecutive frames taken from an input video.
	\textit{A.1}: Dominant object mask generated for segmentation of image \textit{A}.
	\textit{Flow}: Optical flow between images \textit{A} and \textit{B}, represented as an RGB image. This is used for performing automatic labelling of images where pixel movement for every pixel in Image \textit{A} is represented by the calculated optical flow. \textit{A.2}: Training input with a randomly selected reference point (green).
	\textit{B.2}: Horizontally flipped version of image \textit{B}.
	Note the corresponding reference point (green), which is calculated using the optical flow offset.
	The reference point pair is used as a positive example for contrastive loss.
	Additional negative samples for the green reference point in \textit{A.2} are marked in blue.
	\textit{A.3}, \textit{B.3}: Learned descriptors for images \textit{A} and \textit{B}.}
	\label{fig:Optical_Flow}
\end{figure*}

Relative labeling has been in use for various applications. \cite{Wang_Learn} introduced a multi-scale network with triplet sampling algorithm that learns a fine-grained image similarity model directly from images. For image retrieval using deep hashing, \cite{Zhang_bit} trained a deep CNN where discriminative image features and hash functions are simultaneously optimized using max-margin loss on triplet units. However, these methods also require labeled data. \cite{Wang_Gupta} propose a method for image patch detection where they employ relative labeling for data points. Here they try to track patches in videos where two patches connected by a track should have similar visual representation and have closer distance in deep feature space and hence be able to differentiate it from a third patch. In contrast, our method works on pixel level and is thus able to provide fine-grained correspondence. Also, their work does not focus on representing a specific patch distinctively and so it is not clear if a correspondence generated for a pixel is consistent across different scenarios for an object.

Our work makes use of optical flow estimation \cite{Flow_1,Flow_2,Flow_3,FlowNet2_p}, which has made significant progress through deep learning. While these approaches can obviously be used to estimate correspondences for short time frames, our work is applicable for larger geometric transformation, environment with more light variance and occlusion. Of course, a better optical flow estimate during training improves results in our method.

Contrastive loss for dense correspondence has been used by \cite{Schmidt_Sel} where they have used relative labeling to train a fully convolutional network. The idea presented by them is for the descriptor to be encoded with the identity of the point that projects onto the pixel so that it is invariant to lighting, viewpoint, deformation and any other variable other than the identity of the surface that generated the observation. \cite{Florence_Den} train the network for single-object and multi-object descriptors using a modified pixel-wise contrastive loss function which (similar to \cite{Schmidt_Sel}) minimizes the feature distance between the matching pixels and pushes that of non-matching pixels to be at least a configurable threshold away from each other. Their aim is to train a robotic arm to generate training data consisting of objects of interests and then training a FCN network to distinctively identify different parts of the same object and multiple objects. In these methods, an RGB-D video is used and a strong 3D generative model is used to automatically label correspondences. In absence of the depth information, a dense optical flow between subsequent frames of an RGB video can provide the correlation between pixels in different frames. For image segmentation to select the dominant object in the scene, we refer to the solution given by \cite{Chen_Deeplab} and \cite{Jia_Caffe}.

\section{Method}
Our aim is to train a visual descriptor network using an RGB video to get a non-linear function which can translate an RGB image $\mathbb{R}^{W\mathsf{x}H\mathsf{x}3}$ to a descriptor image $\mathbb{R}^{W\mathsf{x}H\mathsf{x}D}$. In the absence of geometry information, two problems need to be solved: The object of interest needs to be segmented for training and pixel correspondences need to be determined. To segment the object of interest from the scene, we refer to the off the shelf solution provided by \cite{Chen_Deeplab}. It provides us with a pre-trained network built on Caffe~\cite{Jia_Caffe} from which masks can be generated for the dominant object in the scene. To obtain pixel correspondences we use a dense optical flow between subsequent frames of an RGB video.

The network architecture and loss function used for training are as explained below: 

\subsection{Network Architecture}
A fully convolution network (FCN) architecture is used where the network has 34 layered ResNet\cite{ResNet} as the encoder block and 6 deconvolutional layers as the decoder block, with ReLU as the activation function. ResNet used is pre-trained on ImageNet \cite{imagenet_cvpr09} data-set. The input has the size $H{\times}W{\times}3$ and the output layer has the size $H{\times}W{\times}D$, where D is the descriptor size. The network also has skip connections to retain information from earlier layers, and L2 normalization at the output.

\subsection{Loss function}
A modified version of pixelwise contrastive loss \cite{Hadsell_Dim}, as has been used by \cite{Florence_Den}, is used for training. The corresponding pixel pair determined using the optical flow is considered as a positive example and the network is trained to reduce the L2 distance between their descriptors. The negative pixel pair samples are selected by randomly sampling the pixels from the foreground object, where the previously generated mask helps in limiting the sampling domain to just the object of interest. The negative samples are expected to be at least $m$ distance away. We thus define the loss function

\begin{multline}
L(A, B, u_a, u_b, O_{A\rightarrow B}) = \\
\left\{
\begin{array}{l@{}l}
D_{AB}(u_a, u_b)^2 & \text{ if } O_{A\rightarrow B}(u_a) = u_b, \\
\\
\max(0, M - D_{AB}(u_a, u_b))^2 & \text{ otherwise. }\\
\end{array}
\right.
\end{multline}
where $O_{A\rightarrow B}$ is the optical flow mapping of pixels from image $A$ to $B$, $M$ is a margin and $D_{AB}(u_a, u_b)$ is the distance metric between the descriptor of image $A$ at pixel $u_a$ and the descriptor of image $B$ at pixel $u_b$ as defined by
\begin{multline}
D_{AB}(u_a, u_b) = ||f_A(u_a) - f_B(u_b) ||_2
\end{multline}
with the computed descriptor images $f_A$ and $f_B$.

\subsection{Training}
For training the descriptor space for a particular object, an RGB video of the object is used.
Masks are generated for each image by the pre-trained DeepLab network~\cite{Chen_Deeplab} to segment the image into the foreground and background. Two subsequent frames are extracted from the video, and dense optical flow is calculated using Farneback's method~\cite{Farnebaeck_opticalflow} after applying the mask (see \cref{fig:Optical_Flow}). We also used the improvements in optical flow estimation using Flownet2~\cite{FlowNet2_p} where we generated the optical flow through pre-trained networks~\cite{FlowNet2_c} for our data-set and observed that an improvement in optical flow determination does indeed help us in a better correspondence estimation.

The pair of frames are passed through the FCN and their outputs along with the optical flow map are used to calculate the pixel-wise contrastive loss. Since there are millions of pixels in an image, 'pixel pair' sampling needs to be done to increase the speed of training. For this, only the pixels on the object of interest are taken into consideration. To select the pixel pair, the pixel movement is measured using the optical flow information. As shown in figure \ref{fig:comp_pixelwise}, once the pixels are sampled from image $A$, corresponding pixels in image $B$ are found using the optical flow map and these become the pixel matches. For each sampled pixel from image $A$, a number of non-match pixels are sampled uniformly at random from the object in image $B$. These create the pixel non-matches. We experimented with the number of positive pixel matches in the range of 1000 to 10000 matches and observed that for us, 2000-3000 positive pixel matches worked the best for us and this was used for the results. For each match 100-150 non-match pairs are considered. The loss function is applied and averaged over the number of matches and non-matches and then the loss is backpropagated through the network.
We performed a grid search to find the parameters for the training. We noticed that the method is not very sensitive to the used hyperparameters. We also noticed that the network trains in about 10 epochs, with loss values stagnating after that point. Data augmentation techniques such as flipping one image in the pair across its horizontal and vertical axis are also applied.

\section{Experiments}

\begin{figure}
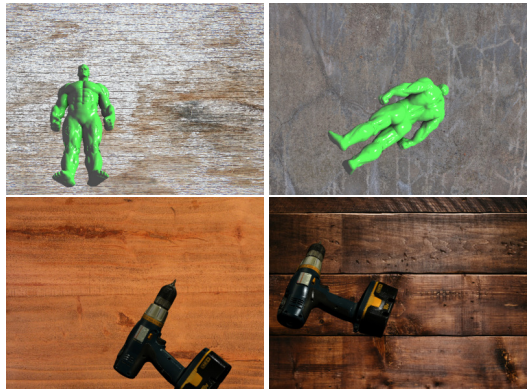

	\centering
	\includegraphics[width=0.45\linewidth]{images/example_images/hulk1.png}
	\includegraphics[width=0.45\linewidth]{images/example_images/hulk2.png}
	\includegraphics[width=0.45\linewidth]{images/example_images/drill1.png}
	\includegraphics[width=0.45\linewidth]{images/example_images/drill2.png}
	\caption{Synthetic images used for experiments. Top row: Hulk figure, Bottom row: Drill. In both cases, we render videos with two different background images.}
	\label{fig:exampleimages}
\end{figure}

For quantitative analysis, we created a rendered video where a solid object is moved - both rotation and translation - in a 3-dimensional space from which we extracted the images to be used for training and evaluation. For all the experiments a drill object is used unless specifed. Different environments under different lighting conditions were created for video rendering. From these videos, frames extracted from the first half of every video is used as train data set and the rest as test data set.

For evaluation, we chose a pixel in one image and compared the pixel representations, which is the output of the network, against the representations of all the pixels in the second image creating an L2 distance matrix. The percentile of total pixels in the matrix that have a distance lesser than the ground truth pixel is determined.

With these intermediate results, we performed the below tests to demonstrate the performance of our method under different conditions:
\begin{enumerate}
	\item We compared percentile of pixels whose feature representation is closer than the representation of the ground truth pixel whose representation is expected to be the closest. This is averaged over an image between consecutive images in a test video. This shows the consistency of the pixel representations of the object for small geometric transformation. For this particular experiment, we used both a drill object dataset and a hulk object dataset (see \cref{fig:exampleimages}). The hulk dataset was trained with the same hyperparameters that were found to be best for drill.\label{Test1}
	\item We selected random images from all the combinations of training and test images with the same and different backgrounds. We performed test similar to \ref{Test1} and results for the train test combination are averaged together. This shows the consistency of the pixel representations of the object for large geometric transformation across various backgrounds. \label{Test2}
	\item We present how the model performs between a pair of images pixel-wise where we plot the percentile of pixels at different error percentiles. \label{Test3}
	\item We show the accuracy of pixel-representation by the presented method for pixels selected through SIFT. This provides a direct comparison of our method with SIFT where we compare the performance of our method on points that are considered ideal for SIFT. \label{Test4}
\end{enumerate}

All these methods are compared against dense SIFT and are plotted together. The method used by \cite{Schmidt_Sel} where they use the 3D generative model for relative labeling is the closest to our method. However, we could not provide an analysis of the results from both the methods as we couldn't make the same setup for quantitative analysis. Also, their method was applicable for RGB-D videos whereas our method as far as we know is new and work on RGB videos and so is different and cannot be compared. We also show visual results from the video recorded and used for training as ground truth is not available for the same.
For test \ref{Test4}, we calculated the significant points for both the images in comparison and determined the significant points for both. We determined SIFT representations for these points and representation for the corresponding pixels were taken from the network output. We then compare the accuracy of the two methods against the ground truth information to check for accuracy.

\section{Evaluation}

\begin{figure}
	\begin{center}
		\includegraphics[width=\linewidth]{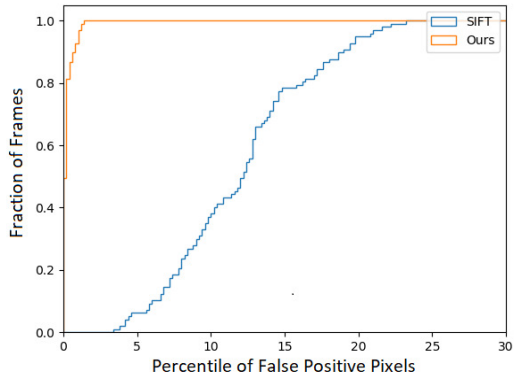}
	\end{center}
	\caption{Cumulative histogram of false positive pixels averaged over a frame between consecutive frames in a video.}
	\label{fig:Video_analysis_graph}
\end{figure}

\begin{figure}[t]
	\begin{center}
		\includegraphics[width=\linewidth]{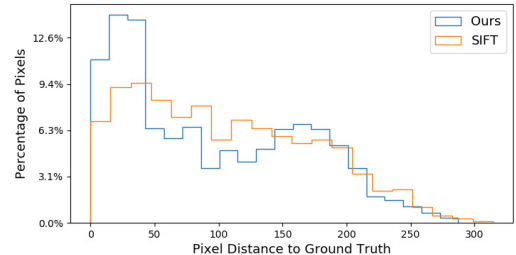}
	\end{center}
	\caption{Correspondence error histogram for keypoints selected through SIFT.
	We show the pixel distance of the estimated corresponding point to the ground truth correspondence for images of size 640*480.
	The presented result considers a different combination of input images based on train and test images with same and different backgrounds.}
	\label{fig:SIFTPreference}
\end{figure}

\begin{table*}
	\centering
	\vspace{1em}
	\caption{
	Quantitative evaluation. We show the percentile of false correspondences
	in each image, averaged over the dataset.
	\textit{Train} refers to images where they have been exposed to the network during training. \textit{Test} refers to images which are previously unseen by the network.
	\textit{Train} and \textit{Test} notations are with respect to our network.
	The results are averaged over images with same and different backgrounds selected at equal proportions.}
	\label{table:imagewisecomparision}
	\begin{tabular}{llrrr}
		\toprule
		Method & Image & Train-Train & Train-Test & Test-Test \\
		\midrule
		SIFT & Same Background & 61.2 & 56.9 & 45.8 \\
		Our Method & Same Background & 3.9 & 8.0 & 3.2 \\
		SIFT & Different Background & 61.6 & 46.0 & 39.2 \\
		Ours Method & Different Background & 4.2 & 1.1 & 1.2 \\
		\bottomrule
	\end{tabular}
\end{table*}

\begin{figure}[t]
	\centering
	\includegraphics[width=\linewidth, height=2.0in]{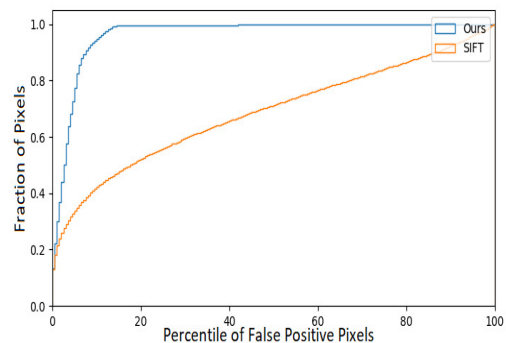}
	\caption{Pixel-wise comparison between the presented method and SIFT. The plot shows the fraction of pixels against percentile of false positives for each pixel, for each of the two approaches.}
	\label{fig:comp_pixelwise}
\end{figure}

\begin{table}
	\centering
	\caption{Image-wise comparison of correspondence results. Here, the error percentile for corresponding images in a video is calculated and is averaged for the whole image and this is repeated for subsequent 100 frame pairs in a video. Images were sampled at 5 frames per second.}
	\label{table:videocorrespondense}
		\begin{tabular}{llrr}
			\toprule
			Method & Object & Mean Error & Std Deviation \\
			\midrule
			SIFT & Drill & 37.0 & 16.4 \\
			Ours & Drill & 0.5 & 0.5 \\
			SIFT & Hulk & 9.5 & 4.8 \\
			Ours & Hulk & 1.6 & 1.9 \\
			\bottomrule
		\end{tabular}
\end{table}

The result for test \ref{Test1} is shown in \cref{table:videocorrespondense}. This demonstrates the overall capability of tracking a pixel across multiple frames in a video and shows how close the representations are in that. As can be seen, our method far outperforms the SIFT method by a large margin when the same test is conducted for the drill object. \Cref{fig:Video_analysis_graph} shows the frame-wise comparison for the same. As can be seen, between most frames, the presented method performs under an error of 1 percentile whereas SIFT results are distributed over the range. We also evaluated the results for the rendered hulk object. The results show a significant difference between drill and hulk object. Our hypothesis for this deviation is that the hulk object has a much better texture than drill and hence SIFT performs much better. We note that training with optimized hyperparameters for the hulk object would likely improve the result.\\
Test \ref{Test2} results are presented in \cref{table:imagewisecomparision}. This test compares the ability of the network to find global representation and representation for pixels not seen before. As can be seen, even upon selecting the previously unseen images and while having different backgrounds between the images, the presented method is able to identify the pixels with similar accuracy proving that the pixel representation found through this method is global. It was observed during the test that the accuracy of prediction depended on the transformation and change in images and not on whether an image and environment has been previously seen by the network proving that the method determines stable features.
Test \ref{Test3} is performed to show the pixel-wise performance of the presented method and the result for the same is shown in \cref{fig:comp_pixelwise}. This shows the overall spread of accuracy of representation, the pixels that are represented well and the outliers. The error percentile for our method performs well for most pixels. As can be seen, for over two-third of the pixels, the error percentile is under 2 percent with close to 15\% pixels under 1 percent error and outperforms the compared method. This is shown in \cref{fig:comp_pixelwise}, it shows the performance of our method when pitted against SIFT. This shows that the representation uniquely identifies the high contrast pixels more easily, better than SIFT.
Test \ref{Test4} is performed to show the capability of the presented method for significant points determined for SIFT algorithm, the results for which are available in \cref{fig:SIFTPreference}. We expected the SIFT to perform well for these significant points since SIFT was developed for this problem. But surprisingly, even for these points, our method performed better than SIFT in most cases as can be seen from the figure.

\section{\uppercase{Conclusions}} \label{sec:conclusion}

\noindent We introduce a method where images from a monocular camera can be used to recognize visual features in an environment which is very important for robots. We present an approach where optical flow calculated using a traditional method such as \cite{Farnebaeck_opticalflow} or using pre-trained networks as given by \cite{FlowNet2_p}, \cite{Flow_3} is used to label the training data collected by the robot enabling it to perform dense correspondence of the surrounding without manual intervention. For recognizing the dominant object, we use a pre-trained network \cite{Chen_Deeplab} which is integrated into this method and can be used without supervision. We present evidence to the global nature of the representation generated by the trained network where the same representation is generated for the object present in different environments and/or at different viewpoints, transformation, and lighting conditions. We also showed that even in cases where a particular image and environment is previously unseen by the network, our network is able to generate stable features with results similar to training images which will allow the robot to explore new environments. We quantitatively compared our approach to a hard-engineered approach scale-invariant feature transform (SIFT) both considering the percentile of false positives pixels averaged over the image and comparison at a pixel level. We show that in both evaluations, our approach far outperforms the existing approach. We also showed our approach's performance with real-world scenarios where the network was trained with a video recorded using a hand-held camera and we were able to show that the network was able to generate visually stable features for the object of interest both in previously seen and unseen environment thereby showing the extension of our method to real-world scenario.\\
Using optical flow for labeling the positive and negative exampled needed for contrastive loss-based training provides us with the flexibility of replacing the optical flow generation method with an improved method when it becomes available and any improvement in optical flow calculation will improve our results further. Similarly, for generating the mask, a more accurate dominant object segmentation will improve the presented results further. The extracted features finds application in various fields of robotics and computer vision such as 3D registration, grasp generation, simultaneous localization and mapping with loop closure detection and following a human being to a destination among others.

\bibliographystyle{apalike}
{\small
\bibliography{paper}}

\begin{thebibliography}{}

\bibitem[Brachmann et~al., 2014]{Brachmann_Learn}
Brachmann, E., Krull, A., Michel, F., Gumhold, S., Shotton, J., and Rother, C.
  (2014).
\newblock {\em Learning 6D Object Pose Estimation Using 3D Object Coordinates},
  volume 8690.

\bibitem[C.~Russell et~al., 2006]{Russel_Usin}
C.~Russell, B., T.~Freeman, W., A.~Efros, A., Sivic, J., and Zisserman, A.
  (2006).
\newblock Using multiple segmentations to discover objects and their extent in
  image collections.
\newblock volume~2, pages 1605 -- 1614.

\bibitem[Chen et~al., 2014]{Chen_Deeplab}
Chen, L.-C., Papandreou, G., Kokkinos, I., Murphy, K., and Yuille, A.~L.
  (2014).
\newblock Semantic image segmentation with deep convolutional nets and fully
  connected crfs.
\newblock {\em CoRR}, abs/1412.7062.

\bibitem[Deng et~al., 2009]{imagenet_cvpr09}
Deng, J., Dong, W., Socher, R., Li, L.-J., Li, K., and Fei-Fei, L. (2009).
\newblock {ImageNet: A Large-Scale Hierarchical Image Database}.
\newblock In {\em CVPR09}.

\bibitem[Farneb{\"a}ck, 2000]{Farnebaeck_opticalflow}
Farneb{\"a}ck, G. (2000).
\newblock Fast and accurate motion estimation using orientation tensors and
  parametric motion models.
\newblock In {\em ICPR}.

\bibitem[Fischer et~al., 2015]{Flow_1}
Fischer, P., Dosovitskiy, A., Ilg, E., Häusser, P., Hazırbaş, C., Golkov,
  V., van~der Smagt, P., Cremers, D., and Brox, T. (2015).
\newblock Flownet: Learning optical flow with convolutional networks.

\bibitem[Florence et~al., 2018]{Florence_Den}
Florence, P.~R., Manuelli, L., and Tedrake, R. (2018).
\newblock Dense object nets: Learning dense visual object descriptors by and
  for robotic manipulation.
\newblock In {\em CoRL}.

\bibitem[G{\"u}ler et~al., 2018]{Guler_Den}
G{\"u}ler, R.~A., Neverova, N., and Kokkinos, I. (2018).
\newblock Densepose: Dense human pose estimation in the wild.
\newblock {\em 2018 IEEE/CVF Conference on Computer Vision and Pattern
  Recognition}, pages 7297--7306.

\bibitem[Hadsell et~al., 2006]{Hadsell_Dim}
Hadsell, R., Chopra, S., and Lecun, Y. (2006).
\newblock Dimensionality reduction by learning an invariant mapping.
\newblock pages 1735 -- 1742.

\bibitem[Hariharan et~al., 2015]{Hariharan_Hyper}
Hariharan, B., Arbelaez, P., Girshick, R., and Malik, J. (2015).
\newblock Hypercolumns for object segmentation and fine-grained localization.
\newblock pages 447--456.

\bibitem[He et~al., 2016]{ResNet}
He, K., Zhang, X., Ren, S., and Sun, J. (2016).
\newblock Deep residual learning for image recognition supplementary materials.

\bibitem[Hinton et~al., 1995]{Hinton_Wake}
Hinton, G.~E., Dayan, P., Frey, B.~J., and Neal, R.~M. (1995).
\newblock The "wake-sleep" algorithm for unsupervised neural networks.
\newblock {\em Science}, 268 5214:1158--61.

\bibitem[Ilg et~al., 2017]{FlowNet2_p}
Ilg, E., Mayer, N., Saikia, T., Keuper, M., Dosovitskiy, A., and Brox, T.
  (2017).
\newblock Flownet 2.0: Evolution of optical flow estimation with deep networks.
\newblock pages 1647--1655.

\bibitem[Janai et~al., 2018]{Flow_2}
Janai, J., Guney, F., Ranjan, A., Black, M., and Geiger, A. (2018).
\newblock {\em Unsupervised Learning of Multi-Frame Optical Flow with
  Occlusions: 15th European Conference, Munich, Germany, September 8-14, 2018,
  Proceedings, Part XVI}, pages 713--731.

\bibitem[Jia et~al., 2014]{Jia_Caffe}
Jia, Y., Shelhamer, E., Donahue, J., Karayev, S., Long, J., Girshick, R.~B.,
  Guadarrama, S., and Darrell, T. (2014).
\newblock Caffe: Convolutional architecture for fast feature embedding.
\newblock In {\em ACM Multimedia}.

\bibitem[Long et~al., 2014]{Long_Fully}
Long, J., Shelhamer, E., and Darrell, T. (2014).
\newblock Fully convolutional networks for semantic segmentation.
\newblock {\em Arxiv}, 79.

\bibitem[Lowe, 2004]{SIFT}
Lowe, D. (2004).
\newblock Distinctive image features from scale-invariant keypoints.
\newblock {\em International Journal of Computer Vision}, 60:91--.

\bibitem[Reda et~al., 2017]{FlowNet2_c}
Reda, F., Pottorff, R., Barker, J., and Catanzaro, B. (2017).
\newblock flownet2-pytorch: Pytorch implementation of flownet 2.0: Evolution of
  optical flow estimation with deep networks.

\bibitem[Schmidt et~al., 2017]{Schmidt_Sel}
Schmidt, T., Newcombe, R.~A., and Fox, D. (2017).
\newblock Self-supervised visual descriptor learning for dense correspondence.
\newblock {\em IEEE Robotics and Automation Letters}, 2:420--427.

\bibitem[Shotton et~al., 2013]{Shotton_Scene}
Shotton, J., Glocker, B., Zach, C., Izadi, S., Criminisi, A., and Fitzgibbon,
  A. (2013).
\newblock Scene coordinate regression forests for camera relocalization in
  rgb-d images.
\newblock pages 2930--2937.

\bibitem[Sivic et~al., 2005]{Sivic_Disc}
Sivic, J., Russell, B., Efros, A., Zisserman, A., and Freeman, W. (2005).
\newblock Discovering objects and their location in images.
\newblock {\em IEEE International Conference on Computer Vision}, pages
  370--377.

\bibitem[Sudderth et~al., 2005]{Sudderth_des}
Sudderth, E.~B., Torralba, A., Freeman, W.~T., and Willsky, A.~S. (2005).
\newblock Describing visual scenes using transformed dirichlet processes.
\newblock In {\em NIPS}.

\bibitem[Sun et~al., 2017]{Flow_3}
Sun, D., Yang, X., Liu, M.-Y., and Kautz, J. (2017).
\newblock Pwc-net: Cnns for optical flow using pyramid, warping, and cost
  volume.

\bibitem[Taylor et~al., 2012]{Taylor_Vit}
Taylor, J., Shotton, J., Sharp, T., and Fitzgibbon, A. (2012).
\newblock The vitruvian manifold: Inferring dense correspondences for one-shot
  human pose estimation.
\newblock volume~10.

\bibitem[Wang et~al., 2014]{Wang_Learn}
Wang, J., song, Y., Leung, T., Rosenberg, C., Wang, J., Philbin, J., Chen, B.,
  and Wu, Y. (2014).
\newblock Learning fine-grained image similarity with deep ranking.
\newblock {\em Proceedings of the IEEE Computer Society Conference on Computer
  Vision and Pattern Recognition}.

\bibitem[Wang and Gupta, 2015]{Wang_Gupta}
Wang, X. and Gupta, A. (2015).
\newblock Unsupervised learning of visual representations using videos.
\newblock {\em 2015 IEEE International Conference on Computer Vision (ICCV)},
  pages 2794--2802.

\bibitem[Zhang et~al., 2015]{Zhang_bit}
Zhang, R., Lin, L., Zhang, R., Zuo, W., and Zhang, L. (2015).
\newblock Bit-scalable deep hashing with regularized similarity learning for
  image retrieval and person re-identification.
\newblock {\em IEEE Transactions on Image Processing}, 24:4766--4779.

\end{thebibliography}

\end{document}